# Precise and Robust Sidewalk Detection: Leveraging Ensemble Learning to Surpass LLM Limitations in Urban Environments

Ibne Farabi Shihab[1] Sudesh Ramesh Bhagat[2] and Anuj Sharma[23]

*Abstract*—This study aims to compare the effectiveness of a robust ensemble model with the state-of-the-art ONEPEACE Large Language Model (LLM) for accurate detection of sidewalks. Accurate sidewalk detection is crucial in improving road safety and urban planning. The study evaluated the model's performance on Cityscapes, Ade20k, and the Boston Dataset. The results showed that the ensemble model performed better than the individual models, achieving mean Intersection Over Union (mIOU) scores of 93.1%, 90.3%, and 90.6% on these datasets under ideal conditions. Additionally, the ensemble model maintained a consistent level of performance even in challenging conditions such as Salt-and-Pepper and Speckle noise, with only a gradual decrease in efficiency observed. On the other hand, the ONE-PEACE LLM performed slightly better than the ensemble model in ideal scenarios but experienced a significant decline in performance under noisy conditions. These findings demonstrate the robustness and reliability of the ensemble model, making it a valuable asset for improving urban infrastructure related to road safety and curb space management. This study contributes positively to the broader context of urban health and mobility.

## I. INTRODUCTION

Autonomous driving has sparked a renaissance in deep learning, transforming the conversation around this technology from a question of possibility to one of optimization [1]. With companies like Tesla unveiling self-driving cars, discussions now focus on autonomy, additivity, LiDAR-based perception of surroundings, and safety ( [2]. The ability of these vehicles, including connected vehicles, to accurately pinpoint their location and distinguish roads from sidewalks is crucial for human well-being.

Detecting sidewalks from images is vital to reducing accidents, ensuring pedestrian safety, and facilitating the smooth operation of autonomous vehicles, even in situations with limited or no internet access. Although classical computer vision techniques such as affine transformation and dynamic contour models have achieved 80% accuracy in sidewalk detection, the stakes are high, and further exploration is necessary [4], [5].

Efficiently detecting sidewalks is crucial for managing curb spaces, enhancing pedestrian safety by separating walkways from vehicular areas, and minimizing the risk of accidents. It also helps regulate traffic flow by designating specific vehicle zones, reducing congestion caused by improper parking. Additionally, these detection systems provide valuable insights for urban planners in developing strategic urban designs and enforcing clear pathways.

Various studies have documented innovative approaches to managing curb space, including automated assessments and integrated management strategies. According to Frackelton et al. [61], automated sidewalk assessment helps with urban design and maintenance by quantitatively evaluating sidewalks. Their research focuses on the condition, accessibility, and continuity of sidewalks. The Southern California Association of Governments also offers an integrated perspective on management, emphasizing priority allocation, dynamic pricing, and digital platforms, which is invaluable for autonomous vehicles navigating urban environments [62].

Hansen et al. discuss adopting sustainable policies and practices in curb space management, highlighting global implementations and policy enforcement challenges [63]. Shaheen et al. explore the ramifications of Transportation Network Companies (TNCs) on curb space, analyzing curb usage strategies that could guide self-driving vehicles' inefficient space utilization [64].

This text examines a framework for curbside management's evolution and future improvement considerations. The framework advises autonomous vehicles on safe and legal stopping or parking zones [65]. Mitman et al. investigate implementing dynamic curb space management. They show how cities adapt curb usage based on time, day, and specific needs. This offers insights into real-time urban curb management that benefit autonomous vehicle operations, as cited in [66].

Traditional object detection methods like YOLO [6], Faster-RCN [7], and Single Shot Multi-box Detector [8] have been extensively employed for detecting various objects within images. However, the intricate nature of urban environments often proves challenging for these methods, which need help to accurately and precisely locate sidewalks. Obstacles such as occlusions, changing light conditions, and the proximity of diverse objects complicate identifying sidewalks with high precision.

To tackle these challenges, we put forth an ensemblebased image segmentation approach that capitalizes on the strengths of multiple segmentation models to accurately detect and delineate sidewalks in a range of urban settings. Image segmentation offers a more refined version of object detection, allowing for precise object localization within images. Our method combines the power of Hierarchical Adaptive Mean Shift (HAMM), DeepLabV3, and YOLACT to create an ensemble model that excels in accuracy, precision, and noise resistance [17], [20], [24], [25]. We also provided the comparison with a state-of-the-art large language model named ONE-PEACE for sidewalk detection [9]. Using ensemble-based image segmentation, we aim to investigate a new, more accurate, and noise-resistant approach to sidewalk detection in street view images. Our method is designed to perform reliably in real-world scenarios, making it ideal for practical applications in urban planning, accessibility analysis, and autonomous vehicle navigation. Our study aims to help anyone interested in contributing to detecting sidewalks or curbs from images. Our paper provides a comprehensive review of related works, including a detailed analysis of the limitations of object detection for sidewalk detection and current LLM models with image segmentation capabilities (Section II). We also propose a new ensemble learning-based method for sidewalk detection based on image segmentation (Section III). Section IV presents our results using ensemblebased learning and the large language models named ONEPEACE. Finally, we discuss our results and offer suggestions for future research in Section V.

---

[1] Ibne Farabi Shihab is with Department of Computer Science, Iowa State University of Science and Technology, Ames, Iowa,50010,United States ishihab@iastate.edu

[2] Anuj Sharma is with Department of Civil Construction and Environmental Engineering,, Iowa State University of Science and Technology, Ames, Iowa,50010,United States anujs@iastate.edu

[3] Sudesh Ramesh Bhagat is with Department of Civil Construction and Environmental Engineering, Iowa State University of Science and Technology, Ames, Iowa,50010,United States bhagat@iastate.edu

The contributions of our work are as follows:

1) This study has comprehensively analyzed the current literature on sidewalk detection and its applicability in urban planning.
2) The study has developed a simple but effective ensemble-based model using state-of-the-art models with different backbones models to surpass the individual models' mIOU score.
3) To the best of our knowledge, this study has introduced LLM for sidewalk detection and compared it with an ensemble-based model for the first time.
4) This work has shown robustness analysis of an ensemble-based model compared to individual models and an LLM with error analysis for future research direction.
5) The study highlights the superior performance of the ensemble model in noisy conditions, offering new strategies for handling such challenges.

## II. BACKGROUND

Object detection concerns discerning an object's category and delineating its bounding box, whereas semantic segmentation involves classifying objects according to their respective categories at the pixel level. Both techniques have potential applications in sidewalk detection; however, this area has seen limited exploration, with most studies concentrating on sidewalk accessibility [3]. Initial attempts to apply object detection methods such as YOLO [6], FasterRCNN [7], and Single Shot Multi-box Detector [8] to sidewalk detection faced challenges due to the need for precise localization of sidewalks rather than bounding boxes.

Two particularly noteworthy studies include that of Chang et al., which employed a deformable part model (DPM) to detect curb ramps [13]. Nonetheless, the DPM proved inadequate for numerous instances of localizing images containing multiple objects. From a machine learning standpoint, Support Vector Machines (SVMs) and Bag of Words have been utilized in image analysis [10], [71], yet their outcomes have been less than promising. However, more recently, deep learning has begun to eclipse other methodologies, garnering significant academic interest in various aerial-based object detection tasks [11], [12]. Sun and Jacob approached the challenge from a distinct angle, electing to focus on identifying what was absent from an image rather than what was present; in their case, this involved detecting missing curb ramps. They integrated a fully convolutional network with a Siamese network and combined machine learning and computer vision techniques to produce a context map. This map accounted for the missing curb ramps, incorporated with object detection to locate the missing regions. Nonetheless, as with the prior study, the 27% recall score leaves room for improvement [3], [27]. Research on sidewalk detection using object detection methods has focused on constructing bounding boxes around sidewalks to ascertain their approximate location. However, this approach can not provide the precision necessary for accurate sidewalk localization, as bounding boxes cannot capture sidewalks' intricate shapes and boundaries. This limitation becomes particularly evident when considering the need for autonomous vehicles to navigate safely and precisely near sidewalks.

In the context of segmentation, Segnet is among the first models relevant to this investigation, employing an encoderdecoder structure for pixel-based segmentation [14]. For sidewalk detection, it hovers around 80% accuracy when using the KITTI dataset [16]. Advanced iterations such as DeepLab, DeepLabv3, and Hierarchical Multi-Scale Attention for Semantic Segmentation (HMASS) have achieved impressive mIOU scores in sidewalk detection [15], [17], [20], [25]. These models have benefited from using backbone networks such as Resnet-50, Xception, Xception-JFT, and Resnet-101, and transfer learning approaches to reduce data and computational resource requirements.

Researchers have contributed significantly to the identification and accessibility enhancement of sidewalks. For instance, the system introduced by Saha and colleagues leverages sensor data combined with machine learning algorithms to assess sidewalk accessibility, providing timely notifications to municipal authorities about potential issues

[55]. Nonetheless, the expansion of this system faces challenges such as limited resources and privacy issues.

In a distinct approach, Hara and colleagues employ image recognition and geographic information systems (GIS) to explore physical attributes of sidewalks like blockages and uneven surfaces, examining their effects on accessibility [56]. Their research aims to facilitate better route planning, especially for self-driving vehicles in urban settings.

Weld and co-researchers delve into deep learning to identify accessibility problems on sidewalks, including barriers and unevenness [57]. They highlight the efficiency of sophisticated algorithms in these scenarios and introduce a specialized dataset for research purposes. The study also points out the necessity for adaptable models to accommodate diverse sidewalk environments.

Through analyzing street view imagery, Chacra and team focus on identifying safety hazards such as cracks and potholes, pivotal for the safety of pedestrians and vehicles alike [58]. The dynamic and constantly evolving urban landscapes pose significant challenges to this method.

Yang and associates developed a system for detecting sidewalks using a lightweight semantic segmentation network to assist visually impaired people by indicating sidewalk boundaries [59]. While this system is geared towards pedestrian use, adapting it for the more dynamic scenarios faced by autonomous vehicles may require extensive modifications.

The discussion wraps up by examining Gamache and colleagues' innovative method, which merges crowdsourcing with machine learning to map out sidewalk accessibility for people with disabilities [60]. This approach identifies issues like irregular pavement and has potential applications in enhancing navigation for autonomous vehicles. Nevertheless, the reliance on crowdsourced information raises questions about data reliability and its universal application.

Ensemble learning methods have been widely employed in machine learning and computer vision tasks to enhance the performance of individual models by leveraging their complementary strengths. These methods combine the predictions of multiple models to form a more accurate and robust final output. Examples of ensemble learning techniques in computer vision include the combination of multiple classifiers for handwritten digit recognition [47], the fusion of different deep learning architectures for image classification [48], [70], and the integration of multiple segmentation models for medical image segmentation [49]. Although recent advances have yielded promising results, sidewalk detection remains complex due to the significant variability in sidewalk structures depending on location. A single model may face difficulties capturing these nuances, as deep models are often susceptible to subtle changes, such as the addition of noise [28]. To overcome these limitations, an ensemble learning-based approach, a strategy extensively employed in machine learning and computer vision tasks to bolster the performance of individual models by capitalizing on their complementary strengths, is necessary. Ensemble methods offer enhanced performance compared to individual learners by aggregating the responses of multiple models using majority voting, thereby generating a more accurate and robust final output [29].

Large Language models (LLM) have gained popularity in various fields since the introduction of BERT in 2019 [53]. Language models have even been applied to object detection and image segmentation, as seen in the Pix2seq paper [44]. Since then, this model has been improved, with Pix2seqD and FitTransformer (FIT) being the latest iteration [51], [52]. However, the multimodal LLM named ONE-PEACE by Wang et al. stands out with its state-of-the-art results on one of the datasets being used [9]. This recent success has prompted further exploration into the application of LLMs [72], [73].

Integrating LLMs into sidewalk detection efforts is motivated by their potential to understand and process complex patterns within vast amounts of data, offering innovative ways to identify and classify objects within images when traditional computer vision techniques may fall short [68]. In addition, the multimodal LLM named ONE-PEACE by Wang et al. stands out with its state-of-the-art results on one of the datasets being used [9]. This recent success has motivated us to explore the application of LLMs on sidewalks. Exploring complementary strategies to enhance model performance is essential to mitigate limitations associated with LLM approaches in object detection and segmentation.

TABLE I: Distribution of datasets

| Dataset Name | Total Images | Training Images | Testing Images |
|---|---|---|---|
| Cityscapes | 2100 | 1680 | 420 |
| Ade20k | 3000 | 2400 | 600 |
| Boston Street | 2000 | 1600 | 400 |

III. MATERIALS AND METHODS

*A. Description of datasets and Preprocessing of Datasets*

This section discusses our chosen data sources, experimental platform, and models employed in our investigation. The efficacy of any machine learning model relies heavily on the quality and relevance of the data it utilizes. For this research, imagery from sources such as Google Earth or street view images is deemed appropriate [30]. However, ensuring that the topography within this data is suitable is essential, as an excessively close view of the images would be counterproductive. Additionally, the requirement for annotated data presents a challenge, as manual annotation is labor-intensive and time-consuming. As a result, we opted for the Cityscapes [18], and Ade20K datasets [19] for our study. However, to diversify our data sources and incorporate additional variations, we also included the dataset utilized by Seong and Jaewan [31] named Boston Dataset. This signifies that our approach will be benchmarked against three distinct datasets.

The Cityscapes dataset, comprising road images from fifty unique cities, offers 5,000 annotated images—sufficient for our project. We selected 2,100 images, allocating 20% for testing and the remaining 80% for training. The Ade20K dataset, encompassing roughly 25,000 images, prompted us to choose 3,000 containing sidewalks to maintain a balanced comparison with the Cityscapes dataset. Adhering to the same pattern as Cityscapes, we designated 80% of the images for training, with the remainder used for testing. Similarly, we extracted 2,000 images from the Boston Street dataset, dividing them into an 80-20 split for training and testing purposes. It is important to note the varied quantities of images taken from each dataset. Our extensive experiments demonstrated that this particular combination of images yielded the best results for our ensemble model. Table I presents a clearer depiction of the percentages and numbers of images from these three datasets. It is important to note that we have used standard data augmentation techniques during training like rotations, flip, and scaling with a probability of 20%, which means out of 5 samples, 1 sample was either flipped, rotated, or scaled to prevent phenomena like overfitting.

### B. metric

The mean Intersection Over Union(mIOU) score is a crucial metric used in computer vision to evaluate the accuracy of image segmentation and object detection models. It calculates model accuracy by averaging the Intersection Over Union (IOU) across all classes. IOU measures the overlap between predicted and ground truth segments relative to their union. mIOU provides a comprehensive and balanced performance assessment, which mitigates class imbalance effects by equally weighing each class's contribution. This metric is essential for applications requiring precise segmentation, as it highlights the model's ability to localize and delineate diverse objects accurately. IOU's significance extends to benchmarking models across tasks like autonomous driving and medical image analysis, guiding toward nuanced detection and segmentation accuracy improvements. Throughout the work, we have used the (mIOU) score to report the score of our models.

### C. Computational Resource

Training deep models is time-consuming and expensive in terms of resources. To overcome these issues, we used the Lamda stack server [33] built with 24 GB Nvidia 3090 graphics and 12 cores. Using this server significantly reduced our experimental time and allowed us to explore different types of models. Pytorch was used as a deep learning library [32].

### D. Architecture Selection and Rationale

According to the preliminary experiment, bounding boxes produced by these models were insufficient for accurate sidewalk detection as they failed to capture the precise shape of sidewalks and were unsuitable for extending to curb detection [Figure 1]. Given the limitations of bounding box-based models, we decided to focus on segmentation models, which offer more accurate object localization. We evaluated various segmentation architectures based on their performance in previous research, suitability for sidewalk detection, tackling different aspects of sidewalk detection, and computational efficiency. After extensive trial and error, while keeping the architectural side in mind, we selected HAMM, DeepLabV3, and YOLACT as our primary models due to their superior performance and diversity in the architectural side in their datasets and respective papers.

After a meticulous selection process, we selected HAMM, DeepLabV3, and YOLACT as the core components of our ensemble model. We conducted several comparative experiments and carefully considered each model's architectural strengths, as well as their relevance to the challenges of sidewalk detection. We started by selecting HAMM for its hierarchical attention mechanisms, adept at isolating pertinent features amidst urban clutter. This makes it invaluable for discerning sidewalk boundaries amidst diverse backgrounds.

We also incorporated DeepLabV3, which uses atrous convolutions and spatial pyramid pooling to address the problem of scale variation and capture intricate details of sidewalks. Lastly, we included YOLACT for its real-time instance segmentation capabilities, which offer the essential balance between speed and accuracy crucial for autonomous vehicles navigating in real-time urban settings. The synergistic integration of these models leverages HAMM's attention-based feature refinement, DeepLabV3's precision in handling scale and detail, and YOLACT's efficiency in processing to create a robust ensemble capable of accurately detecting sidewalks across varied urban landscapes.

Despite its promising performance in previous research, YOLACT struggled with sidewalk detection in our experiments. The primary challenges were the choice of suitable backbone networks, training difficulties, and the quality of results. The segmentations produced by YOLACT were often inaccurate, leading to false positives and negatives. Additionally, the model's sensitivity to image size further complicated the selection of optimal backbone networks. To address these issues, we experimented with various backbone networks, including R-101-FPN, R-50-FPN, and D-53FPN, and tested their performance on different image sizes (400x400, 550x550, and 700x700) [35]. After fine-tuning and selecting the appropriate backbone networks and image sizes, we enhanced YOLACT's performance. We trained the mentioned backbone networks on our datasets

(Cityscapes, Ade20k, Boston Dataset). After exhaustive trial and error, we identified four backbone networks that performed well on our datasets: Resnet-50 for HAMM and Xception, Xception-JFT, and Resnet-101 for DeepLabV3. We found Resnet-50 to be the only suitable backbone for HAMM with our datasets. For YOLACT, we selected R-101-FPN, R-50-FPN, and D53-FPN based on our experiments. Here, R denotes Resnet [37], FPN represents feature pyramid network [38], and D signifies darknet [39]. The numbers 50,101, and 53 indicate the number of layers.

Transfer learning is an efficient way of using machine learning models trained on one task to solve a related task with limited data available for training. This technique helps improve the model's performance and training efficiency by utilizing the knowledge acquired from the first task. The process involves keeping the initial layers of the pretrained model that have learned to recognize general features applicable across tasks while fine-tuning the latter layers to adapt to the new task's specifics. This method reduces the need for extensive computational resources and large datasets, making it an effective approach to accelerate the training process. We utilized transfer learning to fine-tune the final two layers of our chosen models, preserving the pre-trained models' initial weights and minimizing the data and computational resource needs. As depicted in Table III, V, VII our selected datasets demonstrate the effectiveness of this strategy. By combining the strengths of HAMM, DeepLabV3, and YOLACT, we devised an ensemble model capable of accurately detecting and demarcating sidewalks across diverse urban landscapes. A pictorial view of the process is shown in Figure 2. We also tested the above-stated datasets using the LLM model named ONE-PEACE because of its state-of-the-art mIOU score of 63.0 on the ade20k Dataset. The performance of this LLM has been shown in Table VIII.

Incorporating transfer learning to fine-tune our models, we closely followed the protocols for loss functions described in the original papers, adapting them to our unified training approach of 25 epochs across all models. This detailed attention to loss functions is critical for optimizing model performance, particularly when adapting pre-trained models to new tasks like sidewalk detection in diverse urban landscapes.

For the HAMM model, we opted for a combination of cross-entropy loss and dice loss to improve the model's handling of class imbalance, common in urban imagery, where sidewalks make up a small portion of the overall scene. This hybrid loss function ensures that the model learns to classify every pixel accurately and emphasizes correctly identifying less prevalent classes, like sidewalks.

We trained DeepLabV3 with a weighted cross-entropy loss function designed to counter the varying frequencies of different classes within our urban dataset. This loss function aims to mitigate the overshadowing effect of more dominant scene elements by assigning higher weights to rarer classes, such as sidewalks, thereby enhancing the model's sensitivity to critical but less represented features.

YOLACT was optimized with a focal loss function that focuses more on hard-to-classify examples and less on easy negatives, making it particularly effective in urban areas where differentiating between sidewalks and other surfaces can be difficult. The focal loss helps reduce the overwhelming influence of abundant, easy negatives, allowing the model to focus its learning efforts on refining the detection of sidewalks.

Per the guidelines in each model's papers, we maintained a consistent learning rate strategy throughout the training process, with minor adjustments to suit our 25-epoch training regimen. Our careful calibration of loss functions and learning rates ensured that each model could effectively learn from pre-trained weights while being fine-tuned to the specifics of sidewalk detection.

When developing a model, it's important to evaluate its robustness. There are various ways to appraise the robustness of a deep learning model. This study uses the noise perturbation technique, following the approach outlined by Zheng, Song, and Goodfellow [50].

This approach introduces Three different noise levels, with standard deviations of 0.01, 0.05, and 0.1, representing low, mid, and high noise, respectively. These perturbations are applied to produce test datasets consisting of the resulting images. To evaluate the model's performance on these noisy images, the metric of mIOU score is used.

The resulting mIOU scores, obtained under the varied noise levels, are meticulously cataloged. These scores, as documented in Table IX, X, and XI, provide an exhaustive depiction of the model's robustness in the face of varying degrees of noise perturbation with compare to ONE-PEACE. To comply with other works and for ease of comparison, the mIOU score has been used.

This section describes the experimental platform, datasets, transfer learning methodology, backbone networks, and the reasoning behind our architectural choices. We also include a robustness check and the use of ONE-PEACE LLM. These components lay the groundwork for our ensemble model, which we will discuss further in the results section.

## IV. RESULTS

Conducting separate experiments for each dataset is recommended to ensure a thorough analysis of each dataset. This approach allows for a comprehensive evaluation of the ensemble model's robustness.

### A. Experiment on Cityscapes

This study's initial focus was on the Cityscapes dataset, examining various image sizes and an array of backbone networks. The experiments yielded notable results with the Resnet-50 backbone for HAMM, achieving an mIOU score of 89.2% [37]. For DeepLabv3, multiple backbones exhibited promising performance, including Xception with an mIOU score of 87.8%, Xception-JFT with 89.0%, and Resnet-101 with 75.6% [40]. In the case of YOLACT, the top five models best suited for the Cityscapes dataset were identified, with the selection process elaborated in the methodology section. Consequently, nine models were obtained, and their mIOU scores are reported in Table II.

DeepLabV3 Xception model was deliberately excluded to preserve diversity among the three models, as including two DeepLabV3 models would compromise the ensemble's concept. As a result, the mIOU score of 93.1% reinforced the validity of our methodology. This 93.1% signifies the most advanced performance for the Cityscapes dataset featuring sidewalks, as documented in Table III. A visual representation of segmentation following ensemble implementation is depicted in Figure 3, while the original image is displayed in Figure 4.

The objective was to enhance the mIOU scores of the models on the Cityscapes dataset. An ensemble method was employed to achieve this goal, selecting nine models with the highest mIOU scores for further analysis. Examples of these models include YOLACT550 with R-101-FPN backbone (84.3% mIOU score) and YOLACT700 with R-101-FPN backbone (87.2% mIOU score) [37].

The ensemble composition encompassed YOLACT700, HAMM, and DeepLabV3 with Resnet-50. Despite yielding two high mIOU score-based DeepLabV3 models, the

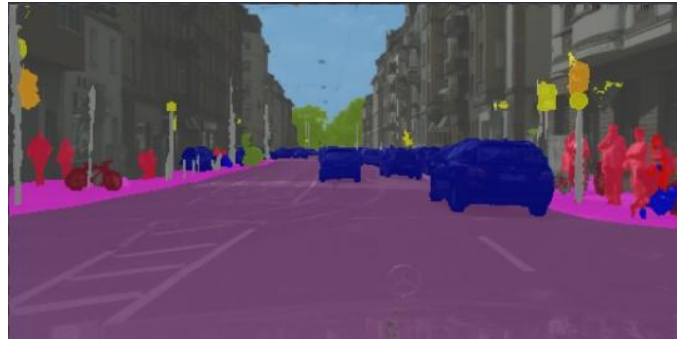
Fig. 3: Segmented image produced by model((Cityscapes))

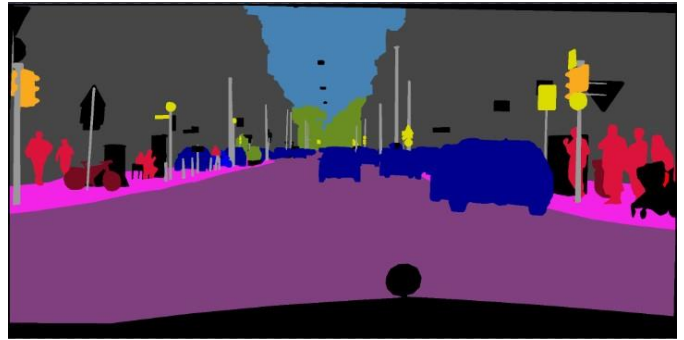
Fig. 2: Pictorial view of work

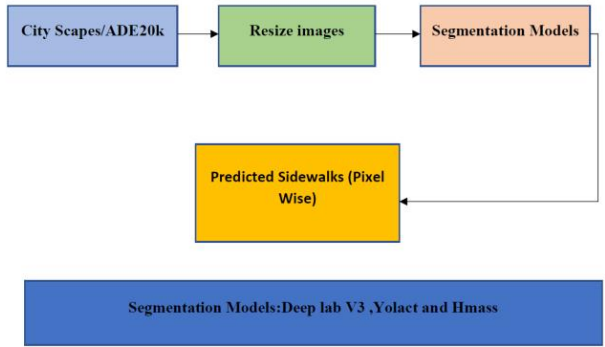

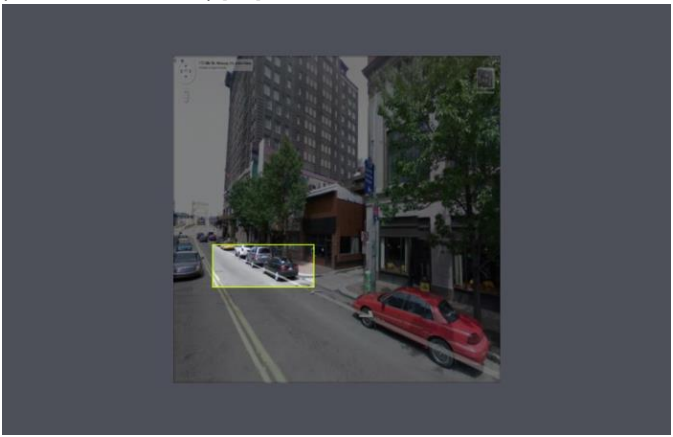
Fig. 1: An example of a bounding box using Faster-RCN from the cityScapes dataset

TABLE II: Performance of Cityscapes dataset on different model

| Network Name (Input Size) | Backbone | mIOU Score (%) |
|---|---|---|
| YOLACT400 | R-101-FPN | 82.2 |
| YOLACT550 | R-101-FPN | 84.3 |
| YOLACT550 | R-50-FPN | 86.5 |
| YOLACT550 | D-53-FPN | 83.4 |
| YOLACT700 | R-101-FPN | 87.2 |
| HAMM | Resnet-50 | 89.2 |
| DeepLabV3 | Xception | 87.8 |
| DeepLabV3 | Xception-JFT | 89.0 |
| DeepLabV3 | Resnet-101 | 75.6 |

TABLE III: Performance of Cityscapes dataset on ensembled model

| Network Name | Backbone | mIOU Score (%) | Ensembled Score |
|---|---|---|---|
| YOLACT700 | R-101-FPN | 87.2 | – |
| HAMM | Resnet-50 | 89.2 | 93.1% |
| DeepLabV3 | Xception-JFT | 89.0 | – |

*B. Experiment on Ade20k*

In this subsection, we scrutinize the performance of the models on the Ade20k dataset, maintaining the same models employed for the Cityscapes dataset in our analysis. For the Ade20k dataset, three models demonstrated the most favorable mIOU scores: YALACT550 with R-50-FPN backbone (85.1% mIOU score), HAMM with Resnet-50 backbone

(88.1% mIOU score), and DeepLabV3 with Xception-JFT backbone (88.1% mIOU score), as documented in Table IV.
Fig. 4: Targeted image from the segmentation(Cityscapes)

An illustration of segmentation post-ensemble implementation can be observed in Figures 5 and 6.

TABLE IV: Performance of Ade20k dataset on different model

| Network Name (Input Size) | Backbone | mIOU Score (%) |
|---|---|---|
| YOLACT400 | R-101-FPN | 81.3 |
| YOLACT550 | R-101-FPN | 84.7 |
| YOLACT550 | R-50-FPN | 85.1 |
| YOLACT550 | D-53-FPN | 84.4 |
| YOLACT700 | R-101-FPN | 82.5 |
| HAMM | Resnet-50 | 88.1 |
| DeepLabV3 | Xception | 87.8 |
| DeepLabV3 | Xception-JFT | 88.1 |
| DeepLabV3 | Resnet-101 | 79.2 |

TABLE V: Performance of Ade20k dataset on ensembled model

| Network Name | Backbone | mIOU Score (%) | Ensembled Score |
|---|---|---|---|
| YOLACT550 | R-50-FPN | 85.1 | – |
| HAMM | Resnet-50 | 88.1 | 90.3% |
| DeepLabV3 | Xception-JFT | 88.1 | – |

*C. Experiment on Boston dataset*

Finally, we applied our nine models to the Boston Dataset and documented the mIOU scores for each model in Table VI. Table VI indicates that YOLACT550 with an R-50-FPN backbone (86.5% mIOU score), HAMM with Resnet-50 backbone (88.5% mIOU score), and DeepLabV3 with Xception-JFT backbone (89.2% mIOU score) exhibited optimal performance in terms of mIOU score. As a result, we selected these three models for the subsequent ensemble round.

Following the methodology of the previous two subsections, we constructed an ensemble of the top-performing models using our voting classifier, attaining a state-of-the-art mIOU score of 90.6% for sidewalk detection on the Boston Dataset. This score can be found in Table VII.

TABLE VI: Performance of Boston dataset on different model

| Network Name (Input Size) | Backbone | mIOU Score (%) |
|---|---|---|
| YOLACT400 | R-101-FPN | 81.2 |
| YOLACT550 | R-101-FPN | 84.8 |
| YOLACT550 | R-50-FPN | 86.5 |
| YOLACT550 | D-53-FPN | 81.7 |
| YOLACT700 | R-101-FPN | 82.9 |
| HAMM | Resnet-50 | 88.5 |
| DeepLabV3 | Xception | 88.3 |
| DeepLabV3 | Xception-JFT | 89.2 |
| DeepLabV3 | Resnet-101 | 80.2 |

TABLE VII: Performance of Boston dataset on ensembled model

| Network Name | Backbone | mIOU Score (%) | Ensembled Score |
|---|---|---|---|
| YOLACT550 | R-50-FPN | 86.5 | – |
| HAMM | Resnet-50 | 88.5 | 90.6% |
| DeepLabV3 | Xception-JFT | 89.2 | – |

*D. Performance of ONE-PEACE LLM on Cityscpaes Dataset, ADE20k Datasets and Boston dataset*

TABLE VIII: Performance of ONE-PEACE LLM

| Dataset Name | mIOU Score (%) |
|---|---|
| CityScapes | 93.8 |
| ADE20k | 91.2 |
| Boston Dataset | 91.4 |

Table VIII provides detailed mIOU scores for the performance of the ONE-PEACE LLM on three datasets: CityScapes, ADE20k, and the Boston Dataset. The results indicate that the ONE-PEACE LLM achieved high mIOU scores of 94.1% on the CityScapes Dataset, 92.5% on the ADE20k Dataset, and 92.8% on the Boston Dataset, which demonstrates its slightly superior (below 1% increase compared to ensembled model) mIOU score compared to our ensemble model's mIOU score across diverse datasets.

*E. Robustness of ensemble models*

Our investigation into the impact of noise on dataset analysis has revealed some interesting findings. Specifically, we found that ensemble models are exceptionally resilient against various noise conditions, unlike standalone models that significantly decline in performance when subjected to noise. Our experimental results, presented in Tables IX, X, and XI, show that our ensemble models can withstand a spectrum of real-life noise scenarios, including Gaussian noise, salt-and-pepper noise, and speckle noise. These noise types are typical of low-light environments, digital transmission errors or corruption, and specialized imaging conditions. Our ensemble approach ensures reliable sidewalk image analysis capabilities, maintaining high mIOU scores even in challenging urban imaging scenarios plagued by foggy conditions, compromised data integrity, and the need for advanced imaging resolution. Remarkably, in environments saturated with high-intensity noise, our ensemble model sustains its performance and surpasses that of individual models, achieving impressive mIOU scores that underscore its superiority and the effectiveness of our tailored approach. This robust performance is consistent across all the datasets. Under noise-inflicted conditions, the ensemble model persistently outperforms the individual models.

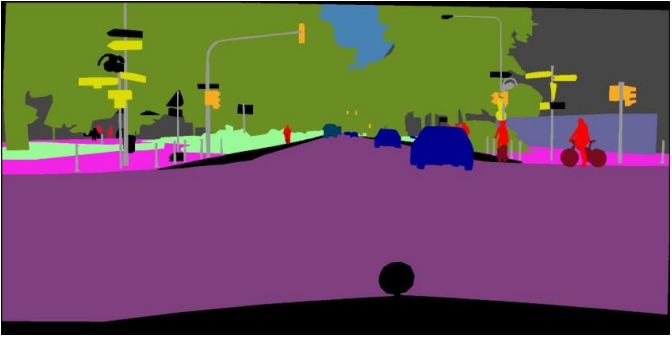

Fig. 5: Segmented image produced by model(Ade20k)

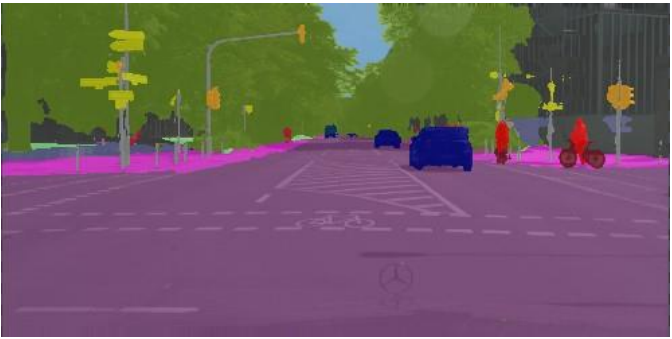

Fig. 6: Targeted image from the segmentation(Ade20k)

Although ONE-PEACE LLM has demonstrated superior performance in normal scenarios (as shown in Table VIII), we have observed that this LLM's performance is more adversely affected by the introduction of noise compared to our proposed ensemble approach (Table IX, X, XI). These empirical observations highlight the strength of ensemble models. The resilience of these models in the presence of noisy data - a common characteristic of real-world data - is particularly impressive. The gradual and controlled decline in accuracy, as opposed to a sharp drop, is a testament to the robustness of ensemble models. Therefore, the robustness of ensemble models presents a reliable option for applications that must work with inherently noisy datasets, even when compared to state-of-the-art LLM models in image segmentation like ONE-PEACE.

*F. Model Performance and Noise Resilience in Diverse Conditions*

The challenges different models face in coping with varying noise levels can be attributed to their unique architectural and operational characteristics. YOLACT, designed mainly for real-time instance segmentation, had difficulty detecting sidewalks as its performance significantly decreased under different noise conditions. This can be linked to its reliance on accurately identifying object boundaries, which can be particularly challenging in the presence of Gaussian noise that blurs these edges and salt-and-pepper noise that introduces stark, misleading contrasts. Furthermore, YOLACT's sensitivity to image size and the need for precise alignment of the backbone network further exacerbate its vulnerability to noise, as these factors depend on the clarity and consistency of visual features within the image. HAMM and DeepLabV3, while generally more resilient than YOLACT, still experienced performance declines as noise levels increased. With its focus on using attention mechanisms for image segmentation, HAMM may struggle in high-noise environments where the relevant features are obscured or distorted, making it challenging to focus on the sidewalk's essential elements. DeepLabV3, known for its effectiveness in semantic segmentation through atrous convolution and spatial pyramid pooling, also experiences a drop in accuracy as noise interferes with the model's ability to interpret spatial relationships and feature variances across the image accurately.

The performance gap between the ensemble models and the ONE-PEACE LLM in noisy conditions highlights the effectiveness of the ensemble. Although ONE-PEACE exhibits exceptional capabilities in clean data scenarios, its performance declines sharply when introducing noise. This is likely due to its design, which may not account for real-world noise's erratic and unpredictable nature as effectively as the ensemble approach. This comparison emphasizes the critical importance of model architecture and training strategy in developing robust image segmentation solutions capable of performing reliably in inherently noisy and complex realworld environments.

We conducted a detailed analysis to compare our ensemble models with the Large Language Model (ONE-PEACE) in various scenarios. We observed that both approaches have their strengths and limitations depending on the data type, noise, and the level of precision required. In scenarios with clean and noise-free data, ONE-PEACE LLM performs exceptionally well due to its advanced pattern recognition capabilities. Its potential for applications where data quality is strictly controlled or high pattern recognition is crucial, such as in laboratory conditions or digital simulations, is impressive. However, the ensemble models perform better in real-world environments where data is often affected by various types and intensities of noise. For example, in lowlight urban environments with Gaussian noise, or scenarios with transmission errors leading to salt-and-pepper noise, the ensemble models show remarkable resilience, maintaining accuracy. The ensemble models can also handle complex imaging techniques susceptible to speckle noise, highlighting its comprehensive design to counteract unpredictable realworld noise. The contrast in performance becomes more evident in highly dynamic or adverse conditions. For instance, rapidly changing weather affects outdoor surveillance or imaging for autonomous navigation. In such cases, the ensemble models exhibit robustness to high-intensity noise, ensuring reliability. Conversely, ONE-PEACE's performance under these conditions is less reliable, indicating its vulnerability to environmental variability and noise. This could limit its utility in applications requiring dependable performance across various operational scenarios.

TABLE IX: Performance of the ensembled model and ONE-PEACE on the Cityscapes dataset under different types and levels of noise

| Noise Type | Noise Level | YOLACT700 | HAMM | DeepLabV3 | Ensemble Score | ONE-PEACE LLM |
|---|---|---|---|---|---|---|
| Gaussian | Low | 85.3 | 88.0 | 87.6 | 91.8 | 91.6 |
|  | Medium | 82.5 | 86.5 | 85.9 | 90.2 | 88.2 |
|  | High | 78.8 | 84.1 | 83.3 | 88.6 | 86.5 |
| Salt and Pepper | Low | 84.5 | 87.5 | 87.2 | 91.6 | 91.2 |
|  | Medium | 81.8 | 85.9 | 85.4 | 89.9 | 88.2 |
|  | High | 77.3 | 83.0 | 82.3 | 88.1 | 85.9 |
| Speckle | Low | 86.0 | 88.5 | 88.3 | 92.0 | 91.8 |
|  | Medium | 83.2 | 87.1 | 86.7 | 90.5 | 88.6 |
|  | High | 79.0 | 84.9 | 84.2 | 89.0 | 86.7 |

TABLE X: Performance of the ensembled model and ONE-PEACE on the Ade20k dataset under different types and levels of noise

| Noise Type | Noise Level | YOLACT550 | HAMM | DeepLabV3 | Ensemble Score | ONE-PEACE LLM |
|---|---|---|---|---|---|---|
| Gaussian | Low | 83.9 | 87.1 | 87.5 | 91.0 | 90.9 |
|  | Medium | 80.7 | 85.6 | 86.2 | 89.5 | 88.3 |
|  | High | 77.2 | 83.8 | 84.6 | 87.9 | 85.1 |
| Salt-and-Pepper | Low | 82.8 | 86.3 | 87.1 | 90.8 | 90.6 |
|  | Medium | 79.5 | 84.9 | 85.7 | 89.2 | 88.1 |
|  | High | 75.8 | 82.5 | 83.0 | 87.4 | 85.0 |
| Speckle | Low | 84.2 | 87.8 | 88.0 | 91.3 | 91.0 |
|  | Medium | 81.0 | 86.0 | 86.5 | 89.7 | 88.2 |
|  | High | 77.5 | 84.2 | 84.9 | 88.1 | 85.9 |

TABLE XI: Performance of the ensembled model and ONE-PEACE on the Boston dataset under different types and levels of noise

| Noise Type | Noise Level | YOLACT550 | HAMM | DeepLabV3 | Ensemble Score | ONE-PEACE LLM |
|---|---|---|---|---|---|---|
| Gaussian | Low | 85.3 | 87.9 | 88.8 | 90.4 | 90.1 |
|  | Medium | 82.0 | 86.7 | 87.6 | 90.0 | 88.8 |
|  | High | 78.5 | 84.5 | 85.9 | 88.3 | 86.1 |
| Salt-and-Pepper | Low | 84.6 | 87.0 | 88.3 | 90.1 | 89.8 |
|  | Medium | 80.9 | 85.3 | 87.1 | 89.7 | 88.4 |
|  | High | 76.7 | 82.8 | 83.6 | 87.9 | 85.0 |
| Speckle | Low | 85.7 | 88.2 | 89.0 | 90.6 | 90.2 |
|  | Medium | 82.4 | 86.8 | 87.8 | 90.2 | 88.8 |
|  | High | 78.9 | 84.7 | 85.1 | 88.5 | 86.0 |

Based on our comparative analysis between the ensemble models and the ONE-PEACE LLM, we have examined the performance dynamics of these models regarding the noise level. When noise levels are low, the ONE-PEACE LLM performs commendably, almost matching the ensemble models across various datasets. This indicates the robustness of the LLM in environments with minimal disturbances, where its pattern recognition capabilities remain largely unaffected. One-PEACE's performance remains highly competitive for applications such as indoor monitoring systems or precisionfocused analyses with low-level background noise.

As the noise intensity increases from low to medium, the gap in performance between the ensemble models and ONE-PEACE widens. The ensemble models' design, which integrates multiple detection strategies, enables them to adapt more smoothly to the incremental increase in noise. This adaptability is especially useful in urban traffic monitoring or outdoor event analysis, where varying degrees of noise are expected but do not reach extreme levels. In these scenarios, the ensemble models' ability to maintain higher accuracy becomes a defining strength, offering more reliable data interpretation.

At high noise levels, the difference in model performance becomes more pronounced. The ensemble models' resilience is highlighted as they deliver robust segmentation results despite the challenging conditions. This resilience is critical for applications in adverse environmental conditions, such as severe weather surveillance or high-density urban imaging, where noise significantly impacts data quality. In contrast, ONE-PEACE's susceptibility to high-intensity noise markedly diminishes its utility, struggling to maintain accuracy as environmental variability and noise intensity surge.

*G. Error Analysis*

An examination of image segmentation of the ensemble models across five generally categorized scenarios revealed varying precision in sidewalk segmentation, identifying key challenges and improvement areas. The analysis underscores the need for model refinement to boost accuracy in sidewalk identification.

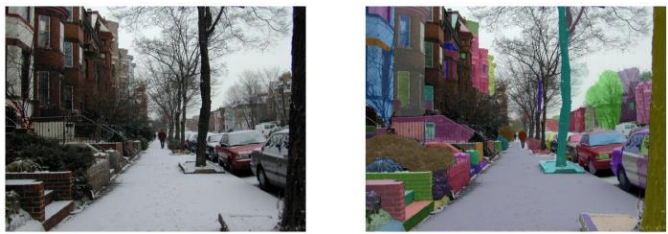

Fig. 7: Exclusion of Tree

Figure 7 shows the model's competency in identifying sidewalk boundaries and its limitation in excluding adjacent tree areas, which is critical for a comprehensive pedestrian pathway analysis. This suggests broader training is necessary to capture the primary pathway features and surroundings.

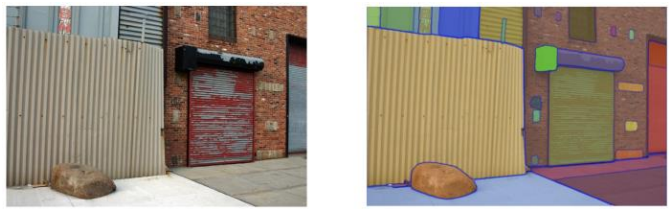

Fig. 8: Example of over-segmentation

In figure 8, the model's over-segmentation, dividing the sidewalk into unnecessary sections, highlights its over sensitivity to minor visual variations. Adjusting the model to appreciate sidewalk continuity could improve its ability to recognize sidewalks as continuous entities.

Figure 9 illustrates under-segmentation, with the model missing a grassy area part of the sidewalk, indicating a training gap in recognizing varied sidewalk compositions. Incorporating a broader range of sidewalk images in training could help the model identify sidewalks in diverse contexts more accurately.

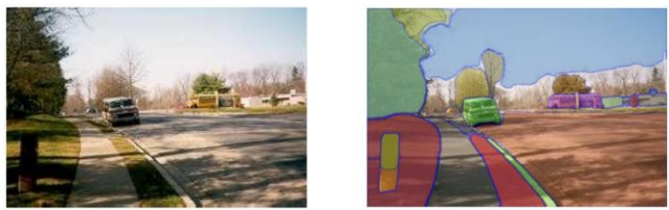

Fig. 9: Example of under-segmentation

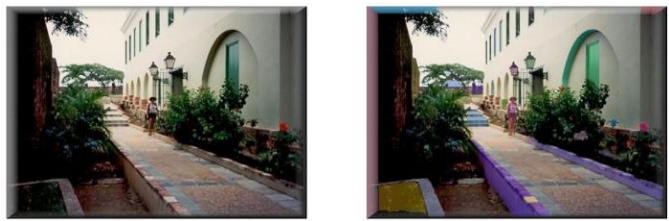

Fig. 10: Exclusion of corner section

The issue in figure 10, where a corner section of the sidewalk is missed, suggests a need for more comprehensive training on complex scenarios and nuanced visual cues to enhance detection accuracy in challenging situations.

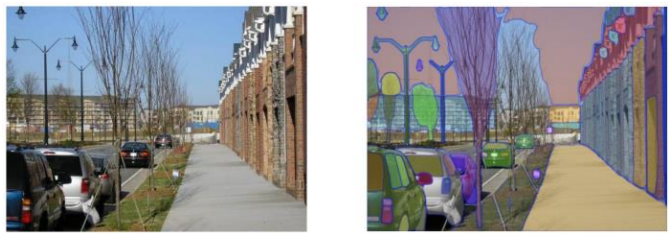

Fig. 11: An ideal example of sidewalk segmentation

Lastly, figure 11 demonstrates ideal sidewalk segmentation, offering a benchmark for model performance. Achieving such precision is crucial for applications like autonomous navigation and medical imaging, emphasizing the value of accurate segmentation.

Our investigation into the impact of noise on dataset analysis has revealed that ensemble models are exceptionally resilient against various noise conditions, unlike standalone models that significantly decline in performance when subjected to noise. Our experimental results in Tables IX, X and XI show that our ensemble models can withstand a spectrum of real-life noise scenarios, including Gaussian noise, saltand-pepper noise, and speckle noise. Lastly, we showed different scenarios where we found our ensemble model underperforming. We, thus, conclude the Results section, highlighting our experimental findings and our capacity to replicate accurate sidewalk detection closely.

## V. Conclusion and Future Work

Our research focused on enhancing sidewalk image analysis to improve autonomous vehicle safety through an ensemble approach. This method demonstrated superior resilience in challenging urban scenarios, including foggy conditions, compromised data integrity, and high-intensity noise environments. Our ensemble model consistently outperformed individual models, maintaining high mIOU scores across various datasets, illustrating its robustness against real-world data challenges.

Comparatively, while the ONE-PEACE LLM showed commendable performance in standard conditions, it was more vulnerable to noise, underlining the strength of our ensemble model in maintaining accuracy in noisy scenarios. This finding underscores the ensemble approach's potential in applications requiring high reliability amidst inherently noisy datasets.

In our pursuit to refine sidewalk detection technologies, we initially explored object detection models like YOLO and Faster RCNN but found them inadequate for capturing the intricate shapes of sidewalks. This led us to semantic image segmentation, where we evaluated several models across the Cityscapes, Ade20K, and Boston datasets to cover a wide range of urban environments. Our analysis identified YOLACT, HAMM, and DeepLabV3 as the top-performing models.

Our study also investigated the impact of image size on model performance, highlighting YOLACT's adaptability to various dimensions. We emphasized the significance of backbone networks and transfer learning in enhancing mIOU scores, which is crucial for our study's accuracy metrics. By developing variations of YOLACT, HAMM, and DeepLabV3 tailored to our datasets, we leveraged transfer learning to fine-tune models, optimizing them for our specific needs. This process involved data augmentation techniques to enrich our training dataset, which comprised thousands of images from the selected urban datasets.

To address computational constraints and variability in mIOU scores among the models tested, we strategically selected three models from different architectures for our ensemble based on their performance synergy. The reason behind choosing three different models from different architecture is that diverse architecture performs better in ensemble setup [69]. This decision was informed by the ensemble's comparative analysis with the state-of-the-art ONE-PEACE model, which, although slightly better in ideal conditions, fell short in noisy environments. Our ensemble model's exceptional performance across all evaluated datasets highlights its robustness and potential to significantly enhance the safety of autonomous vehicles by accurately detecting sidewalks even in challenging conditions.

Exploring future research directions opens up several promising paths. Historically, techniques like segmentation and region proposals have dominated the field, but recent breakthroughs in language modeling present new avenues, especially in areas like machine translation and speech recognition [42], [43]. The potential of language modeling in enhancing image segmentation has begun to surface, offering a novel approach to tackling complex visual tasks. An intriguing possibility lies in applying these advancements to sidewalk detection, leveraging logical semantic segmentation methods proposed by Lai et al. [54]. This could potentially refine the accuracy of detecting sidewalks through a more nuanced understanding of visual data.

The transition towards integrating language models with traditional computer vision techniques acknowledges the limitations of large language models (LLMs) in specific high-precision tasks or when dealing with noisy inputs. This hybrid approach could harness the deep contextual insights of LLMs while supplementing them with the precision of established computer vision methods. Such a combination aims to create a comprehensive system that capitalizes on the strengths of LLMs, addressing their shortcomings through enhanced model accuracy and adaptability in specialized tasks like sidewalk detection. This strategy underscores a pivotal shift towards a more integrated, holistic approach in the field, promising significant improvements in how machines perceive and interpret complex visual environments.

Another avenue to consider is the development of new backbone networks. Backbone networks are predominantly limited to Resnet, Darknet, VGG, or attention mechanismbased networks [38], [40], [45], [46]. It is high time to advance these networks further. Attention mechanisms have undoubtedly proven invaluable in recent years. However, attention must be distributed across various image regions, as multiple objects or objects may possess several noteworthy features. ResNet has adopted a similar approach and demonstrated its potential as a top-tier backbone network [46].

Improving the data quality is the other key to improving object detection models. A closer examination of the data reveals that most come from the same geographical locations, except for the Cityscapes dataset. It's important to note that high-resolution data is necessary for optimal model performance, even though some models can function in lowlight conditions. However, using low-accuracy models can jeopardize safety-critical scenarios, as they may not perform well under challenging conditions. While the Cityscapes dataset includes data from 50 different cities, its usefulness is limited by poor data quality. Poor lighting conditions in images make distinguishing between roads and sidewalks difficult, as demonstrated in a sample image in the experimental platform section. This confusion can cause the model to perform poorly, resulting in a lower score. Lastly, the robustness concept can be refined following the work done in connected -vehicles [?].

In conclusion, our research on sidewalk detection technologies aims to improve the safety of autonomous vehicles. We analyzed different methodologies, data handling, and architecture to create an ensemble model. Our study advances navigation systems and deepens our understanding of object detection models. The results of our work have the potential to enhance road safety, promote pedestrian mobility, and guide urban planning. We hope to pave the way for future research in autonomous vehicle safety, contributing to reliable navigation systems. We aim to foster synergy between technology and urban planning to enhance safety and accessibility worldwide.

## ACKNOWLEDGMENT

The authors would like to thank detectron2 for giving access to the pretrained models, which helped us eliminate annotating.